\documentclass[final,3p,times,twocolumn]{elsarticle}
\usepackage{hyperref}

\journal{Neural Networks}

\pdfoutput=1

\usepackage{graphicx}

\usepackage{amssymb, amsmath}

\usepackage[font={footnotesize},compatibility=false]{caption}
\usepackage[font={footnotesize}]{subcaption}

\usepackage{relsize}

\usepackage{widetext}

\newcommand{\unit}[1]{\,\mathrm{#1}}

\newcommand{\diff}[2]{\frac{\mathrm{d}#1}{\mathrm{d}#2}}

\newcommand{\vek}[1]{%
\ifcat\noexpand#1\relax 
    \boldsymbol{#1}     
  \else                 %
    \mathbf{#1}         
  \fi                   %
}

\newcommand{\uvek}[1]{\hat{\vek{#1}}}

\newcommand{\grad}[1]{\vek{\nabla}_{#1}\,}

\newcommand{\sign}{\mathrm{sign}}

\begin{document}

\begin{frontmatter}

\title{Learning optimal wavelet bases using a neural network approach}

\author{Andreas S{\o}gaard}
\address{School of Physics and Astronomy,\\University of Edinburgh}
\ead{andreas.sogaard@ed.ac.uk}

\begin{abstract}
	A novel method for learning optimal, orthonormal wavelet bases for representing $1$- and $2$D signals, based on parallels between the wavelet transform and fully connected artificial neural networks, is described. The structural similarities between these two concepts are reviewed and combined to a ``wavenet'', allowing for the direct learning of optimal wavelet filter coefficient through stochastic gradient descent with back-propagation over ensembles of training inputs, where conditions on the filter coefficients for constituting orthonormal wavelet bases are cast as quadratic regularisations terms. We describe the practical implementation of this method, and study its performance for a few toy examples. It is shown that an optimal solutions are found, even in a high-dimensional search space, and the implications of the result are discussed.
\end{abstract}


\end{frontmatter}

\section{Introduction} \label{sec:introduction}

\!\footnote{Sections~\ref{sec:introduction} and \ref{sec:theory} contain overlaps with \cite{MScThesis}.}The Fourier transform has proved an indispensable tool within the natural sciences, allowing for the study of frequency information of functions and for the efficient representation of signals exhibiting angular structure. However, the Fourier transform is limited by being global: each frequency component carries no information about its spatial localisation; information which might be valuable. Multiresolution, and in particular wavelet, analysis has been developed, in part, to address this limitation, representing a function at various levels of resolution, or at different frequency scales, while retaining information about position-space localisation. This encoding uses the fact that due to their smaller wavelengths, high-frequency components may be localised more precisely than their low-frequency counterparts.

The wavelet decomposition expresses any given signal in terms of a ``family'' of orthonormal basis functions \cite{Mortlet1982a, Mortlet1982b}, efficiently encoding frequency-position information. Several different such wavelet families exist, both for continuous and discrete input, but these are generally quite difficult to construct exactly as they don't possess closed-form representations. Furthermore, the best basis function for any given problem depends on the class of signal, choosing the best among existing functional families is hard and likely sub-optimal, and constructing new bases is non-trivial, as mentioned above. Therefore, we present a practical, efficient method for directly \emph{learning} the best wavelet bases, according to some optimality criterion, by exploiting the intimate relationship between neural networks and the wavelet transform.

Such a method could have potential uses e.g.~in areas utilising time-series data and imaging, for instance --- but not limited to --- EEG, speech recognition, seismographic studies, financial markets as well as image compression, feature extraction, and de-noising. However, as is shown in Section~\ref{sec:example}, the areas to which such an approach can be applied are quite varied. 

In Section~\ref{sec:introduction-previous-work} we review some of the work previously done along these lines. In Section~\ref{sec:theory} we briefly describe wavelet analyses, neural networks, as well as their structural similarity and how they can be combined. In Section~\ref{sec:MeasuringSparsity} we discuss metrics appropriate for measuring the quality of a certain wavelet basis. In Section~\ref{sec:LearningProcedure} we describe the actual algorithm for learning optimal wavelet bases. Section~\ref{sec:implementation} describes the practical implementation and, finally, Section~\ref{sec:example} provides an example use case from high-energy physics.

\section{Previous work} \label{sec:introduction-previous-work}

A typical approach \cite{Mojsilovic2000, Qureshi2008, Pont2011} when faced with the task of choosing a wavelet basis in which to represent some class of signals, is to select one among an existing set wavelet families, which is deemed suitable to the particular use case based on some measure of fitness. This might lead to sub-optimal results, as mentioned above, since limiting the search to a few dozen pre-exiting wavelet families will likely result in inefficient encoding or representation of (possibly subtle) structure particular, or unique, to the problem at hand. To address this shortcoming, considerable effort has already gone into the question of the existence and construction of optimal wavelet bases. 

Ref.~\cite{Thielemann2006} describes a method for constructing optimally matched wavelets, i.e.~wavelet bases matching a prescribed pattern as closely as possible, through lifting \cite{Sweldens1997}. However, the proposed method is somewhat arduous and relies on the specification of a pattern to which to match, requiring considerable and somewhat artificial preprocessing.\footnote{``It is difficult to find a problem our method can be applied to without major modifications.'' \cite[p.~125]{Thielemann2006}.} This is not necessarily possible, let alone easy, for many use cases as well as for the study of more general \emph{classes} of inputs rather than single examples. In a similar vein, Ref.~\cite{Hurley2007} provides a method for unconstrained optimisation of a wavelet basis with respect to a sparsity measure using lifting, but has the same limitations as Ref.~\cite{Thielemann2006}.

Refs.~\cite{Zhuang1994, Zhuang1996} provide theoretical arguments for the existence of optimal wavelet bases as well as an algorithm for constructing such a basis for single $1$- or $2$D inputs, based on gradient descent. However, results are only presented for low-order wavelet bases, the implementation of orthonormality constraints is not discussed, and the question of generalisation from single inputs to classes of inputs is not addressed. In addition, the optimal filter coefficients referenced in \cite[Table~$1$]{Zhuang1996} do not satisfy the explicit conditions (C2), (C3), and (C4) for orthonormality in Section~\ref{subsec:theory-wavelet} below. These constraints are violated at the $1\%$-level, which also corresponds roughly to the relative angular deviation of the reported optimal basis from the Daubechies \citep{Daubechies1992} basis of similar order. 

Finally, Refs.~\cite{Tewfik1992, Gopinath1994} provide a comprehensive prescription for designing wavelets that optimally represent signals, or classes of signals, at some fixed scale $J$. However, the results are quite cumbersome, are based on a number of assumptions regarding the characteristics of the input signal(s), and relate only to the question of optimal representation at fixed scales.

This indicates that, although the question of constructing optimal wavelet bases has been given substantial consideration, and clear developments have been made already, a general approach to easily learning discrete, demonstrably orthonormal wavelet bases of arbitrary structure and complexity, optimised over \emph{classes} of input has yet to be developed and implemented for practically arbitrary choice of optimality metric. This is what is done below.

\section{Theoretical concepts} \label{sec:theory}

In this section, we briefly review some of the underlying aspects of wavelet analysis, Section~\ref{subsec:theory-wavelet}, and neural networks, Section~\ref{subsec:theory-neuralnetowrk}, upon which the learning algorithm is based. In Section~\ref{subsec:theory-combining} we discuss the parallels between the two concepts, and how these can be used to directly learn optimal wavelet bases.

\subsection{Wavelet} \label{subsec:theory-wavelet}

Numerous excellent references explain multiresolution analysis and the wavelet transform in depth, so the present text will focus on the discrete class of wavelet transforms, formulated in the language of matrix algebra as it relates directly to the task at hand. For a more complete review, see e.g.~\citep{MScThesis} or \cite{Daubechies1992,Mallat1989,Meyer1992,Ripples,WaveletsInEngineering}.

In the parlance of matrix algebra, the simplest possible input signal $\vek{f} \in \mathbb{R}^{N}$ is a column vector
	\begin{equation}
		\vek{f} = \begin{bmatrix}
			\vek{f}[0] \\
			\vek{f}[1] \\
			\vdots \\
			\vek{f}[2^{M}-2] \\
			\vek{f}[2^{M}-1] 
		\end{bmatrix}
	\end{equation}
and the dyadic structure of the wavelet transform means that $N$ must be \emph{radix} $2$, i.e.~$N=2^{M}$ for some $M \in \mathbb{N}_{0}$.\footnote{Although the results below are also applicable to $2$D, i.e.~matrix, input, cf.~Section~\ref{sec:example}.} The forward wavelet transform is then performed by the iterative application of \emph{low-} and \emph{high-pass filters}. Let $L(\vek{f})$ denote the low-pass filtering of input $\vek{f}$, the $i$'th entry of which is then given by the convolution
	\begin{equation} \label{eq:low-pass-filter-entry}
		L(\vek{f})[i] = \sum_{k = 0}^{2^{M}-1} a[k] \vek{f}[i + 	N / 2 - k], \quad  i \in [0, 2^{M-1} - 1]
	\end{equation}
assuming periodicity, such that $\vek{f}[-1] = \vek{f}[N-1]$, etc. The low-pass filter, $a$, is represented as a row vector of length $N_{\mathrm{filt}}$, with $N_{\mathrm{filt}}$ even, and its entries are called the \emph{filter coefficients}, $\{a\}$.

The convolution yielding each entry $i$ in $L(\vek{f})$ can be seen as a matrix inner product of $\vek{f}$ with a row matrix of the form
	\begin{equation}
		\begin{bmatrix}
			& \cdots & 0 & a[N-1] & \cdots & a[1] & a[0] & 0 & \cdots &
		\end{bmatrix}
	\end{equation}
Since this is true for each entry, the full low-pass filter may be represented as a $(2^{M-1} \times 2^{M} ) \cdot (2^{M} \times 1)$ matrix inner product:
	\begin{equation}
		L(\vek{f}) = \vek{L}_{M-1} \, \vek{f}
	\end{equation}
where, for each low-pass operation, the matrix operator is written as
\begin{widetext}
	\begin{equation} \label{eq:lowpass-filter-matrix-explicit}
		\vek{L}_{m} = \underbrace{\begin{bmatrix}
			~~\ddots & \ddots & \ddots & \ddots &        &        &        &        &        &          \\
			~~\cdots & a[N-1] & \cdots & a[1]   & a[0]   & 0      & 0      & 0      & 0      & \cdots~~ \\
			~~\cdots & 0      & 0      & a[N-1] & \cdots & a[1]   & a[0]   & 0      & 0      & \cdots~~ \\
			~~\cdots & 0      & 0      & 0      & 0      & a[N-1] & \cdots & a[1]   & a[0]   & \cdots~~ \\
			         &        &        &        &        &        & \ddots & \ddots & \ddots & \ddots~~ 
		\end{bmatrix}}_{\mathlarger{2^{m+1}}} \left\} \begin{matrix} ~ \\[2.5mm] ~ \\ ~ 2^{m} \\ ~ \\[2.5mm] ~ \end{matrix} \right.
	\end{equation}
\end{widetext}

In complete analogy to Eq.~\eqref{eq:lowpass-filter-matrix-explicit}, a high-pass filter matrix $\vek{H}_{m}$ can be expressed as a $2^{m} \times 2^{m+1}$ matrix parametrised in the same way by coefficients $\{b\}$, which we choose \cite{Daubechies1992} to relate to $\{a\}$ by 
	\begin{equation} \label{eq:wavelet-coefficients-b}
		b_{k} = (-1)^{k} \, a_{N_{\mathrm{filt}} - 1 - k} \quad \text{for} \; k \in [0, N_{\mathrm{filt}}-1]
	\end{equation}
The means that, given wavelet coefficients $\{a\}$, we have specified the full wavelet transform in terms of repeated application of matrix operators $\vek{L}_{m}$ and $\vek{H}_{m}$. The filter coefficients will therefore serve as our parametrisation of any given wavelet basis.

At each step in the transform, the power of $2$ that gives the current length of the (partially transformed) input, $n = 2^{m}$, is referred to as the \emph{frequency scale}, $m$. Large frequency scales $m$ correspond to large input arrays, which are able encode more granular, and therefore more high-frequency, information than for small $m$. 

\begin{figure*}
	\centering%
	\begin{subfigure}[b]{0.33\textwidth}
		\centering%
		\captionsetup{justification=centering}%
		\includegraphics[width=0.9\textwidth]{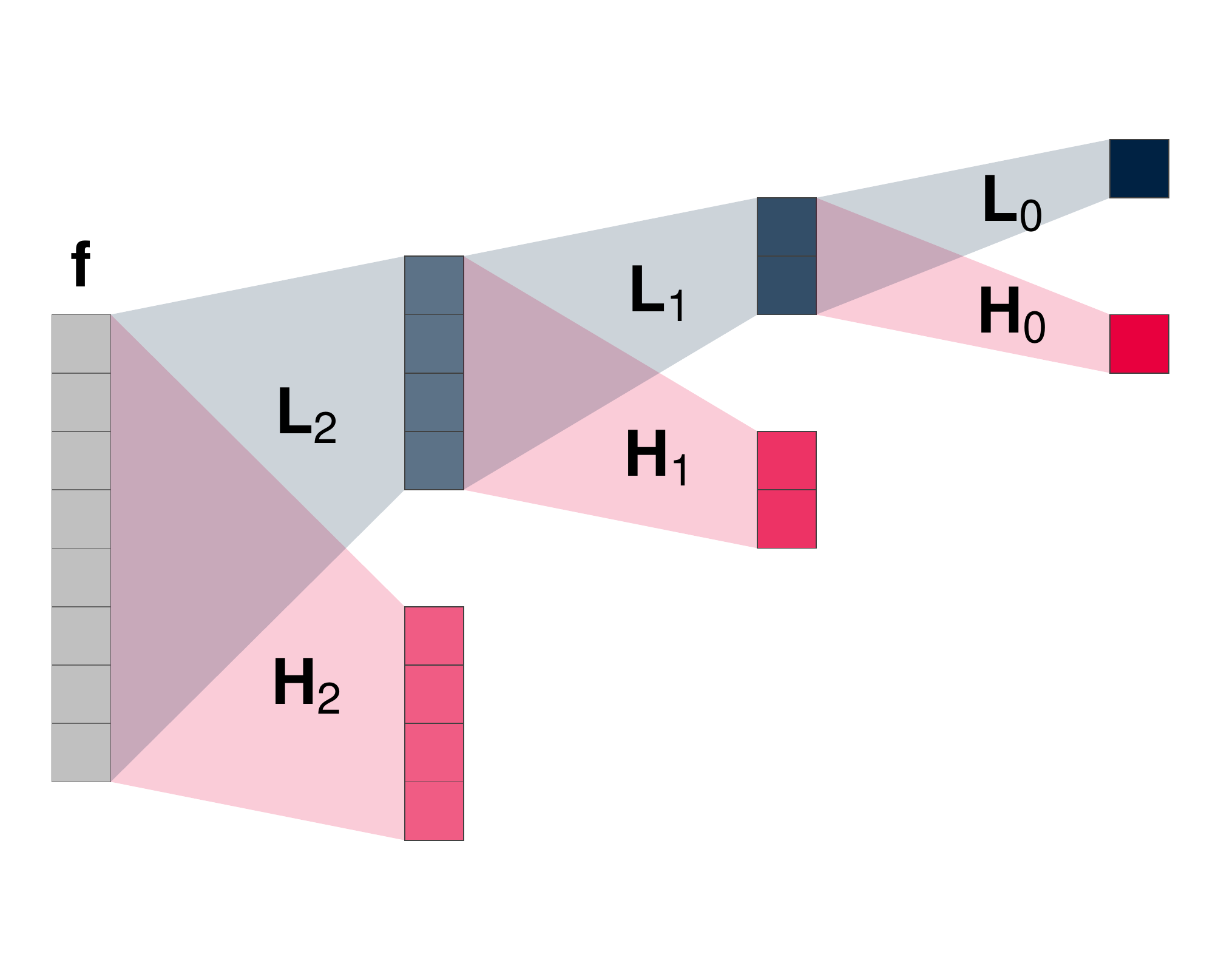}
		\caption{Wavelet}
		\label{subfig:Architectures_Wavelet}
	\end{subfigure}%
	\begin{subfigure}[b]{0.33\textwidth}
		\centering%
		\captionsetup{justification=centering}%
		\includegraphics[width=0.9\textwidth]{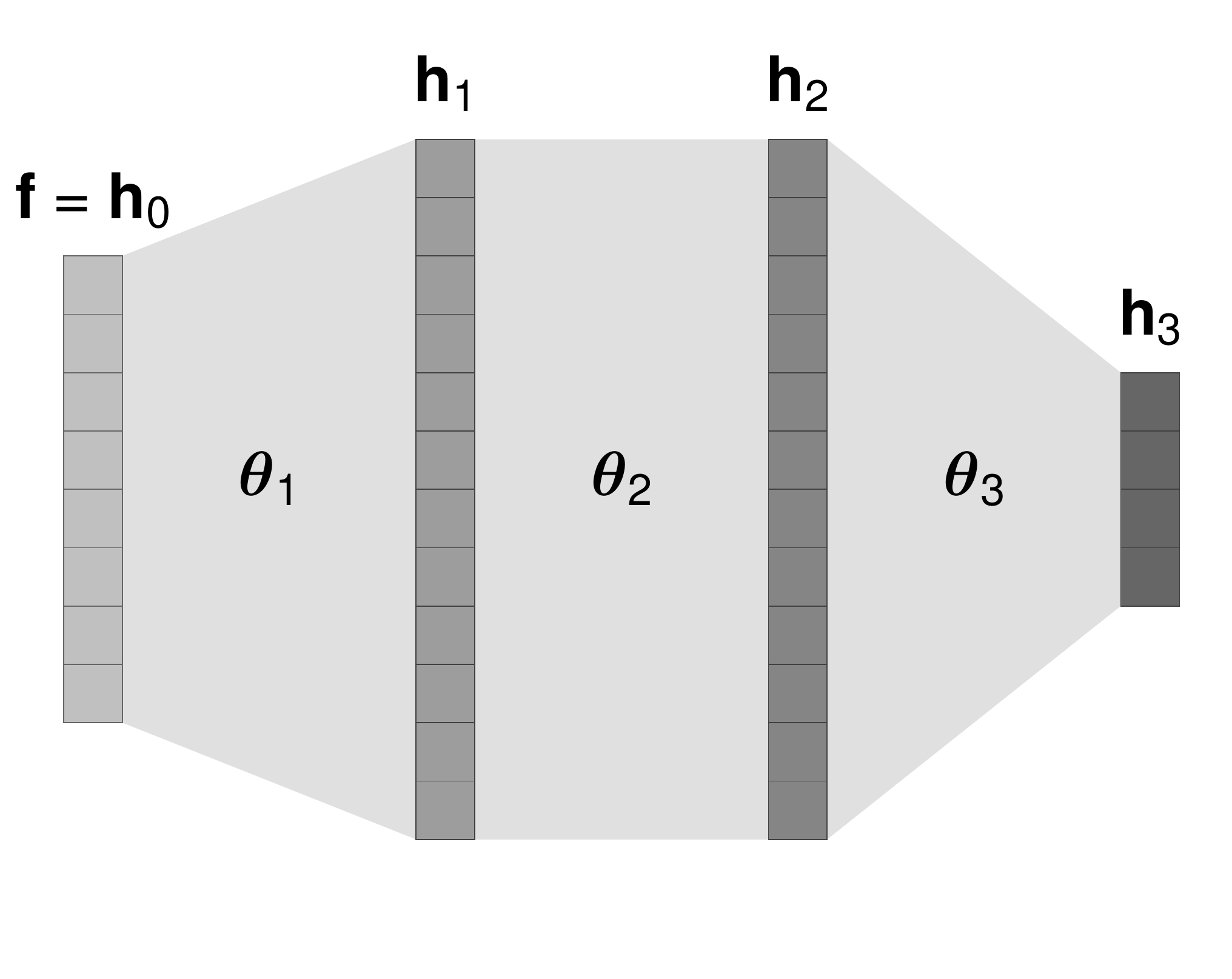}
		\caption{Neural network}
		\label{subfig:Architectures_NN}
	\end{subfigure}%
	\begin{subfigure}[b]{0.33\textwidth}
		\centering%
		\captionsetup{justification=centering}%
		\includegraphics[width=0.9\textwidth]{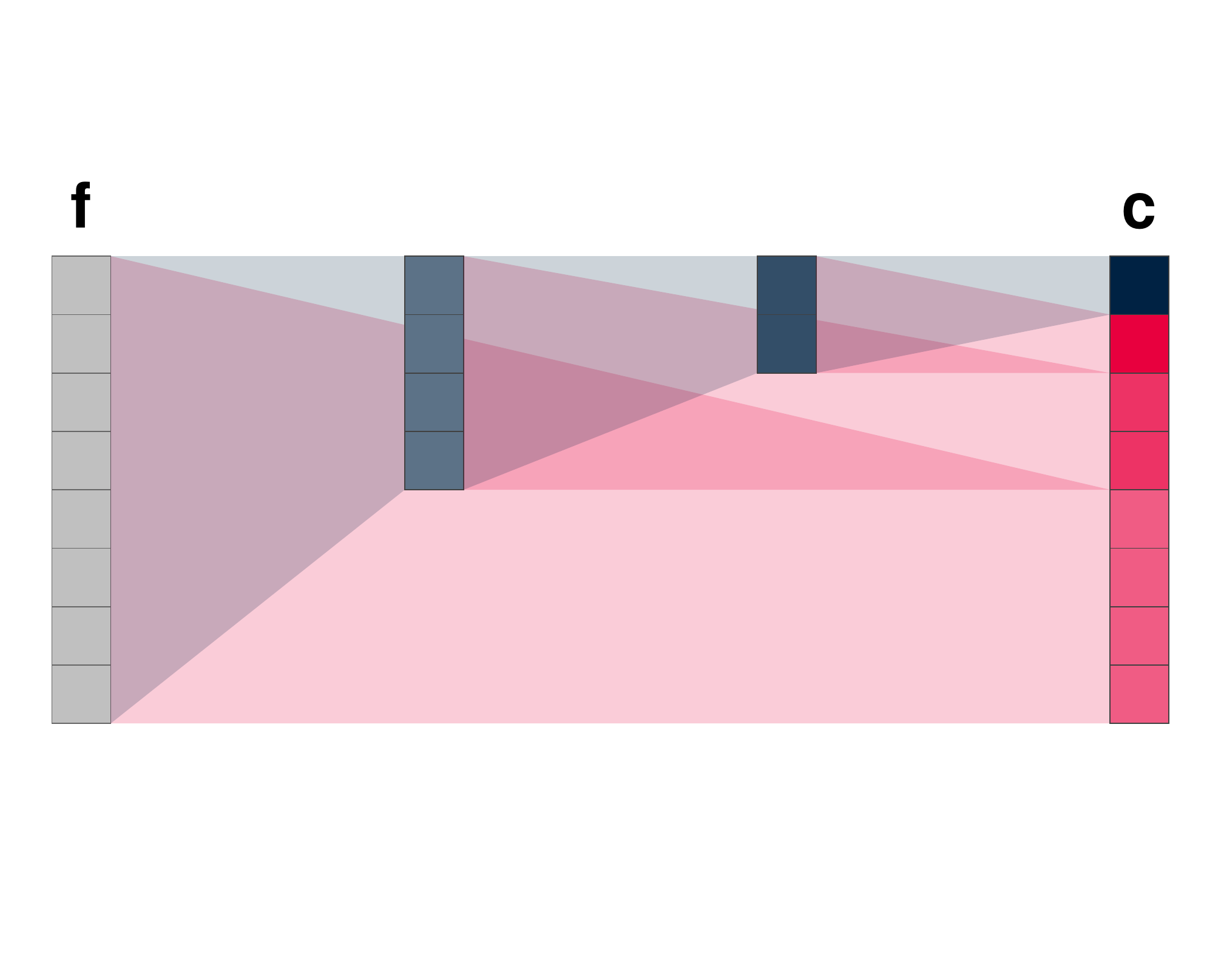}
		\caption{``Wavenet''}
		\label{subfig:Architectures_WaveletNN}
	\end{subfigure}%
	\caption{Schematic representations of the difference in architecture for (a) standard wavelet transforms, (b) fully connected neural networks, and (c) the wavelet transform formulated as a neural network, here called ``wavenet''. Individual squares indicate elements in layers, i.e.~entries in column vectors. Shaded areas indicate filter- or weight matrices, where red/blue represent high-/low-pass filters.}
	\label{fig:Architectures}
\end{figure*}

As the name implies, the low-pass filter acts as a spatial sub-sampling of the input from frequency scale $m$ to $m-1$, averaging out the frequency information at scale $m$ in the process. Similarly, the high-pass filter encodes the frequency information at scale $m$; the information which is lost in the low-pass filtering. After each step, another pass of high- and low-pass filters are applied to the sub-sampled, low-pass filtered input. This procedure is repeated from frequency scale $M$ to $0$. At each step, the high-pass filter encodes the frequency information specific to the current frequency scale. This is illustrated in Figure~\ref{subfig:Architectures_Wavelet}. 

The coefficients obtained through successive convolution of the signal with the high- and low-pass filters, i.e.~the right-most layers in Figure~\ref{subfig:Architectures_Wavelet}, collectively encode the same information as the position-space input $\vek{f}$, but in the basis of wavelet functions. These are called the \emph{wavelet coefficients} $\{c\}$. Given such a set of wavelet coefficients, the inverse transform can be perform by retracing the steps of the forward transform. Letting $\vek{f}_{m}$ denote the input signal low-pass filtered down to scale $m$, with $\vek{f}_{M} \equiv \vek{f}$ the inverse transform proceeds as
	\begin{subequations}
	\begin{align}
		\vek{f}_{0} & = [c_{0}] \\
		\vek{f}_{1} & = \vek{L}_{0}^{\mathrm{T}}  \vek{f}_{0} + \vek{H}_{0}^{\mathrm{T}}  [c_{1}]\\
		\vek{f}_{2} & = \vek{L}_{1}^{\mathrm{T}}  \vek{f}_{1} + \vek{H}_{1}^{\mathrm{T}}  [c_{2} \;\, c_{3}]\\
		& \;\, \vdots \nonumber \\
		\vek{f} \equiv \vek{f}_{M} & = \vek{L}_{M-1}^{\mathrm{T}}  \vek{f}_{M-1} + \vek{H}_{M-1}^{\mathrm{T}}  [\;c_{2^{M-1}} \; \cdots \; c_{2^{M}-1}\;]
	\end{align}
	\end{subequations}
In this way it is seen that $c_{0}$ encodes the average information content in the input signal $\vek{f}$, and that $c_{i>0}$ dyadically encode the frequency information at larger and larger scales $m$. The explicit wavelet basis function corresponding to each wavelet coefficient can be found by setting $c = [\;\cdots \;\, 0 \;\,1\;\,0\;\,\cdots\;]$ and studying the resulting, reconstructed position-space signal $\hat{\vek{f}}$ at some suitable largest scale $M$.

The filter coefficients $\{a\}$ completely specify the wavelet transform and -basis, but they are not completely free parameters, however. Instead, they must satisfy a number of explicit conditions in order to corresponds to an orthonormal wavelet basis. These conditions \citep{Pollock} are as follows:

In order to satisfy the \emph{dilation equation}, the filter coefficients $\{a\}$  must satisfy 
	\begin{align*}
		\sum_{k} a_{k} & = \sqrt{2} \tag{C1}
	\end{align*}
		
In order to ensure orthonormality of the \emph{scaling}- and wavelet functions, the coefficients $\{a\}$ and $\{b\}$ must satisfy
	\begin{align*}
		\sum_{k} a_{k}a_{k+2m} & = \delta_{m,0} \quad \forall \; m \in \mathbb{Z} \tag{C2}
	\end{align*}
and
	\begin{align*}
		\sum_{k} b_{k}b_{k+2m} & = \delta_{m,0} \quad \forall \; m \in \mathbb{Z} \tag{C3}
	\end{align*}
	where the condition for $m=0$ is trivially fulfilled from (C$2$) through Eq.~\eqref{eq:wavelet-coefficients-b}.
		
To ensure that the corresponding wavelets have zero area, i.e.~encode only frequency information, we require
	\begin{align*}
		\sum_{k} b_{k} & = 0 \tag{C4}
	\end{align*}
		
Finally, to ensure orthogonality of scaling and wavelet functions, we must have
	\begin{align*}
		\sum_{k} a_{k} b_{k+2m} & = 0 \quad \forall \; m \in \mathbb{Z} \tag{C5}
	\end{align*}
where condition (C5) is automatically satisfied through Eq.~\eqref{eq:wavelet-coefficients-b}.

Conditions (C$1$--$5$) then collectively ensure that the filter coefficients $\{a\}$ (and $\{b\}$) yield a wavelet analysis in terms of orthonormal basis functions. As we parametrise our basis uniquely in terms of filter coefficients $\{a\}$, since $\{b\}$ are fixed through Eq.~\eqref{eq:wavelet-coefficients-b}, we will need to explicitly ensure that these conditions are met. The method for doing this is described in Section~\ref{subsec:theory-combining}.

\subsection{Neural network}	 \label{subsec:theory-neuralnetowrk}

Since (artificial) neural networks have become ubiquitous within most areas of the physical sciences, we will only briefly review the central concepts as they relate to the rest of this discussion. A comprehensive introduction can be found e.g.~in Ref.~\cite{Bishop1995}.

Neural networks can be seen general mappings $f:\mathbb{R}^{n} \to \mathbb{R}^{m}$, which can approximate any function, provided sufficient capacity. In the simplest case, such networks are constructed sequentially, where the input vector $\vek{f} = \vek{h}_{0} \in \mathbb{R}^{N_{0}}$ is transformed through the inner product with a weight matrix $\vek{\theta}_{1}$, the output of which is a \emph{hidden layer} $\vek{h}_{1} \in \mathbb{R}^{N_{1}}$, and so forth, until the output layer $\vek{h}_{l} \in \mathbb{R}^{N_{l}}$ is reached. The configuration of a given neural network, in terms of number of layers and their respective sizes, is called the network \emph{architecture}. In addition to the transfer matrices $\vek{\theta}_{i}$, the layers may be equipped with \emph{bias nodes}, providing the opportunity for an offset, as well as non-linear \emph{activation functions}. A schematic representation of one such network, without bias nodes and non-linearities, is shown in Figure~\ref{subfig:Architectures_NN}.

The neural network can then be trained on a set of training examples, $\{(\vek{f}_{i}, \vek{y}_{i})\}$, where the task of network usually is to output a vector $\hat{\vek{y}}_{i}$ trying to predict $\vek{y}_{i}$ given $\vek{f}_{i}$. The quality of the prediction is quantified by the \emph{cost} or \emph{objective function} $\mathcal{J}(\vek{y}, \hat{\vek{y}})$. The central idea is then to take the error of any given preduction $\hat{\vek{y}}_{i}$, given by the derivative of the cost function with respect to the prediction at the current value, and \emph{back-propagate} it through the network, performing the inverse operation of the forward pass at each layer. In this way, the gradient of the cost function $\mathcal{J}$ with respect to each entry in the network's weight matrices $(\vek{\theta}_{i})_{jk}$ is computed. Using \emph{stochastic gradient descent}, for each training example one performs small update steps of the weight matrix entries along these error gradients, which is then expected to produce slightly better performance of the network with respect to the task specified by the cost function.

One challenge posed by such a fully connected network is the shear multiplicity of weights for just a few layers of moderate sizes. Such a large number of free parameters can make the network prone to over-fitting, which can be mitigated e.g.~by $L_{2}$ weight \emph{regularisation}, where a regularisation term $\mathcal{R}(\{\vek{\theta}\})$ is added to the cost function, with a multiplier $\lambda$ controlling the trade-off between the two contributions.

\subsection{Combining concepts} \label{subsec:theory-combining}

The crucial step is then to recognise the deep parallels between these two constructs. We can cast the discrete wavelet transform as an $\mathbb{R}^{N} \to \mathbb{R}^{N}$ neural network with a fully-connected, deep, non-sequential, dyadic architecture without bias-units and with linear (i.e.~no) activations. A schematic representation of this setup, here called a ``wavenet'', is shown in Figure~\ref{subfig:Architectures_WaveletNN}. This is done by identifying the neural network transfer matrices with the low- and high-pass filter operators in the matrix formulation of the wavelet transform, cf.~Eq.~\eqref{eq:lowpass-filter-matrix-explicit}. The forward wavelet transform then corresponds to the neural network mapping, and the output vector of the neural network is exactly the wavelet coefficients of the input with respect to the basis prescribed by $\{a\}$. 

If we can formulate an objective function $\mathcal{J}$ for the wavelet coefficients, i.e.~the output of the ``wavenet'', this means that we can utilise the parallel with neural networks and employ back-propagation to gradually update the weight matrix entries, i.e.~the filter coefficients $\{a\}$, in order to improve our wavelet basis with respect to this metric. Therefore, choosing a fixed filter length $|\{a\}| = N_{\mathrm{filt}}$, and parametrising the ``wavenet'' in terms of $\{a\}$, we are able to directly \emph{learn} the wavelet basis which is optimal according to some task $\mathcal{J}$.

Interestingly, and unlike some of the approaches mentioned in Section~\ref{sec:introduction-previous-work}, a neural network approach naturally accommodates \emph{classes} of inputs, in addition to single examples. That is, one can train repeatedly on a single example and learn a basis which optimally represents this particular signal in some way, cf.~e.g.~\cite{Thielemann2006}. However, the use of stochastic gradient descent is naturally suited for fitting the weight matrices to ensembles of training examples, which in many cases is much more meaningful and useful, cf.~Section~\ref{sec:example}.

\begin{figure*}
	\centering%
	\begin{subfigure}[b]{0.5\textwidth}
		\centering%
		\captionsetup{justification=centering}%
		\includegraphics[width=0.9\textwidth]{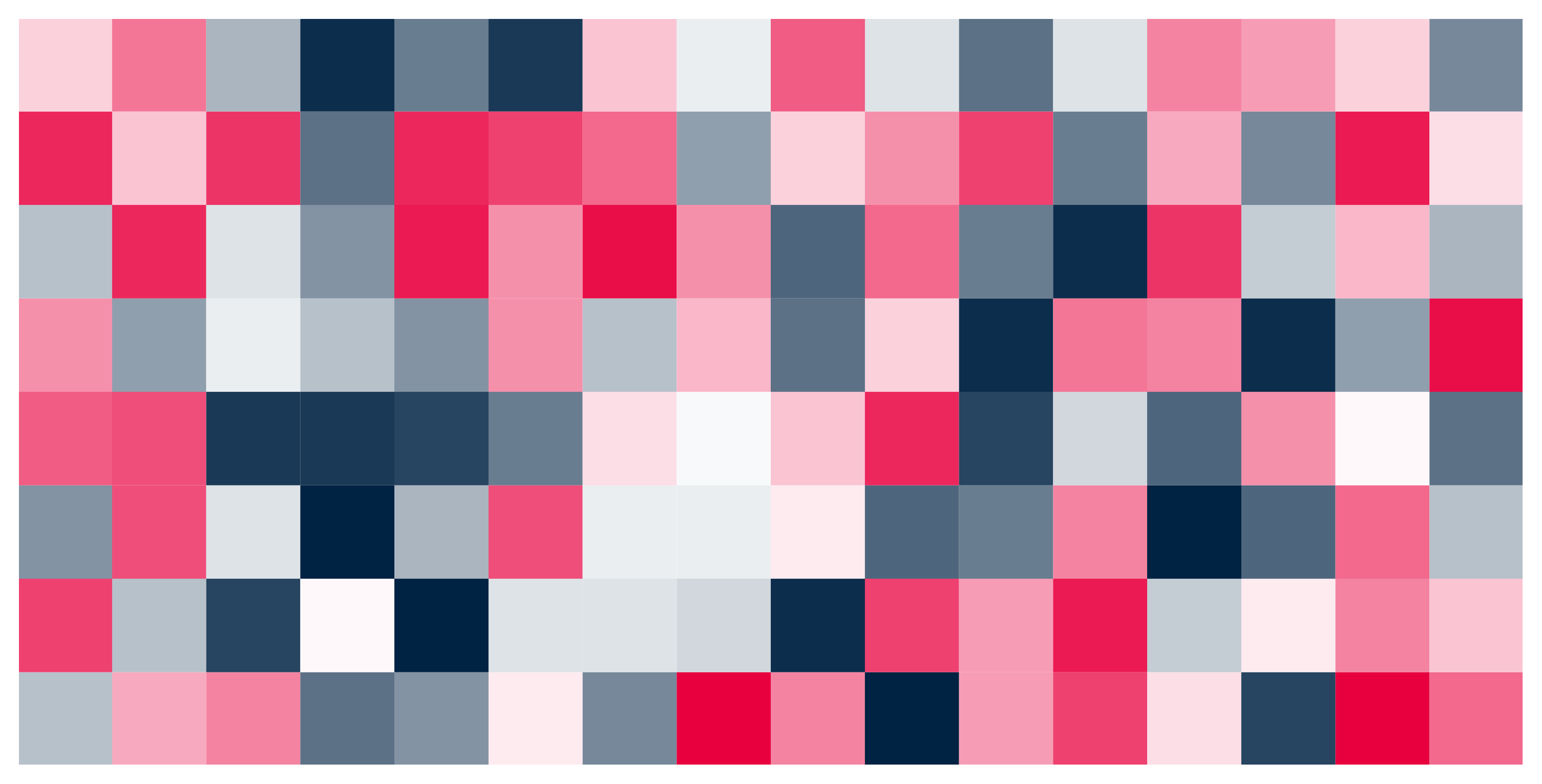}
		\vspace*{1em}%
		\caption{Neural network}
		\label{subfig:TransferMatrix_NN}
	\end{subfigure}%
	\begin{subfigure}[b]{0.5\textwidth}
		\centering%
		\captionsetup{justification=centering}%
		\includegraphics[width=0.9\textwidth]{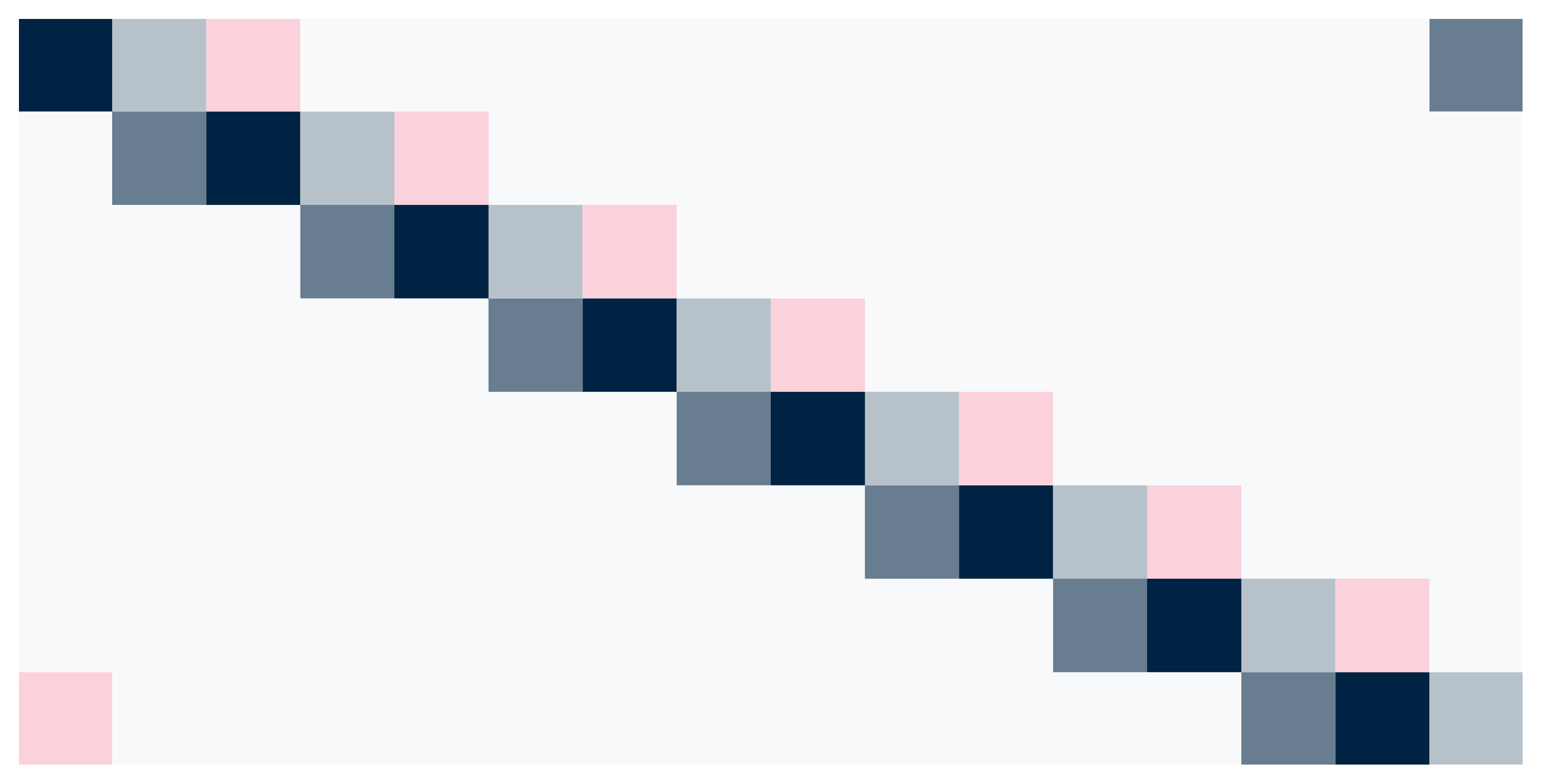}
		\vspace*{1em}%
		\caption{Wavelet}
		\label{subfig:TransferMatrix_Wavelet}
	\end{subfigure}%
	\caption{Schematic representation of the entries in a $8 \times 16$ (a) transfer matrix in a unconstrained, fully connected neural network and (b) a corresponding filter operator in a wavelet transform with $N_{\text{filt}} = 4$ filter coefficients. Note that the entries in each row of the wavelet matrix operator are identical, and simply shifted by integer multiples of $2$, cf.~Eq.~\eqref{eq:low-pass-filter-entry}, such that the number of free parameters is only $N_{\text{filt}}$.}
	\label{fig:TransferMatrices}
\end{figure*}

Another key observation is that while the entries in a standard neural network wight matrix are free parameters, the weights in the ``wavenet'' are highly constrained, since they must correspond to the low- and high-pass filters of the wavelet transform. For instance, a neural network like the one in Figure~\ref{subfig:Architectures_WaveletNN}, mapping $\mathbb{R}^{8} \to \mathbb{R}^{8}$ will have $84$ free parameters in the standard treatment. However, identifying each of the $6$ weight matrices with the wavelet filter operators, this number is reduced to $N_{\mathrm{filt}}$, which can be as low as $2$. This is schematically shown in Figure~\ref{fig:TransferMatrices}. For inputs of ``realistic'' sizes, i.e.~$|\vek{f}| = N \gtrsim 64$ this reduction is exponentially greater, leading to a significant reduction of complexity.

Finally, we note that the filter coefficients need to conform with conditions (C$1$--$5$), cf.~Section~\ref{subsec:theory-wavelet} above, in order to correspond to an orthonormal wavelet basis. This can be solved by noting that all conditions (C$1$--$5$) are differentiable with respect to $\{a\}$, which means that we can cast these conditions in the form of quadratic regularisation terms, $\mathcal{R}_{i}$, which can then be added to the cost function with some multiplier $\lambda$, in analogy to standard $L_{2}$ weight regularisation. The multiplier $\lambda$ then controls the trade-off between the possibly competing objectives of optimising $\mathcal{J}$ and ensuring fulfillment of conditions (C$1$--$5$). In principle, this means that for finite $\lambda$ any learned filter configuration $\{a\}$ might violate these conditions to order $1/\lambda$, and might therefore strictly be taken to constitute a ``pseudo-orthonormal'' basis. This will, however, have little impact in practical application, where one can simply choose a value of $\lambda$ sufficiently high that $\mathcal{O}(1/\lambda)$ is within the tolerances of the use case at hand.

\section{Measuring optimality} \label{sec:MeasuringSparsity}

The choice of objective function defines the sense in which the basis learned through the method outlined in Section~\ref{subsec:theory-combining} will be optimal. This also affords the user a certain degree of freedom in defining the measure of optimality, the only condition being that the objective function be differentiable with respect to the wavelet coefficients $\{c\}$.\footnote{Possibly except for a finite number of points.}

In this example we choose \emph{sparsity}, i.e.~the ability of a certain basis to efficiently encode the information contained in a given signal, as our measure of optimality. From the point of view of compression, sparsity is clearly a useful metric, in that it measures the amount of information that can be stored with a within certain amount of space/memory. From the point of view of representation, sparsity is likely also a meaningful objective, since a basis which efficiently represents the defining features of a (class of) signal(s) will also lead the signal(s) to be sparse in this basis.

Based on \cite{Hurley2009}, we choose the \emph{Gini coefficient} $\mathcal{G}(\,\cdot\,)$ as our metric for the sparsity of a set of wavelets coefficients $\{c\}$, 	
	\begin{align} \label{eq:GiniCoefficient}
		\mathcal{G}(\{c\}) = \frac{ \sum_{i = 0}^{N_{c} - 1} (2 i - N_{c} - 1) |\overline{c}_{i}| }{ N_{c}  \sum_{i = 0}^{N_{c} - 1} |\overline{c}_{i}| } \equiv \frac{f(\{c\})}{g(\{c\})}
	\end{align}
for wavelet coefficients $\{\overline{c}\}$ sorted by ascending absolute value, i.e.~$|\overline{c}_{i}| \leq |\overline{c}_{i+1}|$ for all $i$. Here $N_{c} \equiv |\{c\}|$ is the number of wavelet coefficients.

A Gini coefficient of $1$ indicates a completely unequal, and therefore maximally sparse, distribution, i.e.~the case in which only one coefficient has non-zero value, and therefore carries all of the information content in the signal. Conversely, a Gini coefficient of $0$ indicates a completely equal distribution, i.e.~each coefficient has exactly the same (absolute) value, and therefore all carry exactly the same amount of information content.

Having settled on a choice of objective function, we now proceed to describing the details of the learning procedure itself. We stress that the results of the following sections should generalise to other reasonable choices of objectives, which may be chosen based on the particular use case at hand.

\section{Learning procedure} \label{sec:LearningProcedure}

As noted above, the full objective function for the optimisation problem is given as the sum of a sparsity term $\mathcal{S}(\{c\})$ and a regularisation term $\mathcal{R}(\{a\})$, the relative contribution of the latter controlled by the regularisation constant $\lambda$, i.e.
	\begin{align} \label{eq:combined-cost-function}
		\mathcal{J}(\{c\}, \{a\}) = \mathcal{S}(\{c\}) + \lambda \, \mathcal{R}(\{a\})
	\end{align}
where $\{c\}$ is the set of wavelet coefficients for a given training example and $\{a\}$ is the current set of filter coefficients. The $\mathcal{R}$-term ensures that the filter coefficient configuration $\{a\}$ does indeed correspond to a wavelet basis as defined by conditions (C$1$--$5$) above; the $\mathcal{S}$-term measures the quality of a given wavelet basis according to the chosen fitness measure. The learning task then consists of optimising the filter coefficients according to this combined objective function, i.e.~finding a filter coefficient configuration, in an $N_{\mathrm{filt}}$-dimensional parameter space, which minimises $\mathcal{J}$. 
The procedure for computing a filter coefficient gradient for each of the two terms is outlined below.

\subsection{Sparsity term}
Based on the discussion in Section \ref{sec:MeasuringSparsity}, we have chosen the Gini coefficient $\mathcal{G}(\,\cdot\,)$ as defined in Eq.~\eqref{eq:GiniCoefficient} as our measure of the sparsity of any given set of wavelet coefficients $\{c\}$. The sparsity term in the objective function is chosen to be
	\begin{align} \label{eq:SparsityTerm}
		\mathcal{S}(\{c\}) = 1 - \mathcal{G}(\{c\}) 
	\end{align}
This definition means that low values of $\mathcal{S}(\{c\})$ correspond to greater degree of sparsity, such that that minimising this objective function term increases the degree of sparsity.

In order to utilise stochastic gradient descent with back-propagation, the objective function needs to be differentiable in the values of the output nodes, i.e.~the wavelet coefficients. Since the sparsity term is the only term which depends on the wavelet coefficients, particular care needs to be afforded here. The sparsity term is seen to be differentiable everywhere except for a finite number of points where $\overline{c}_{i} = 0$. In these cases the derivative is taken to be zero, which is meaningful considering the chosen optimisation objective: coefficients of value zero will, assuming at least one non-zero coefficient exists, contribute maximally to the sparsity of the set as a whole. Therefore we don't want these coefficients to change, and the corresponding gradient \emph{should} be zero.\footnote{Cases with all zero-valued coefficients are ill-defined but also practically irrelevant.}

Therefore, assuming $\overline{c}_{i} \neq 0$, the derivative of the sparsity term is given by (suppressing the arguments of the objective function terms for brevity)
	\begin{align} \label{eq:SparsityTermDerivative}
		\grad{|\overline{c}|} \mathcal{S} & \equiv \uvek{e}_{i} \diff{\mathcal{S}}{|\overline{c}_{i}|} =  \uvek{e}_{i} \diff{}{|\overline{c}_{i}|} (1 - \mathcal{G}) \nonumber \\
		& = - \grad{|\overline{c}|} \mathcal{G} \nonumber \\ 
		& = - \frac{ \grad{|\overline{c}|} f \cdot g - f \cdot \grad{|\overline{c}|} g}{g^{2}}   
	\end{align}
where 
	\begin{align}
		\grad{|\overline{c}|} f & = \uvek{e}_{i} \diff{}{|\overline{c}_{i}|} \left( \sum_{k = 0}^{N_{c} - 1} (2 k - N_{c} - 1) |\overline{c}_{k}| \right) \nonumber \\ 
		& = ( 2i - N_{c} - 1 )  \, \uvek{e}_{i} 
\intertext{and} 
		\grad{|\overline{c}|} g & = \uvek{e}_{i} \diff{}{|\overline{c}_{i}|} \left(  N_{c} \sum_{k = 0}^{N_{c} - 1} |\overline{c}_{k}| \right) = N_{c} \,  \uvek{e}_{i} 
	\end{align}
for $f$ and $g$ defined in Eq.~\eqref{eq:GiniCoefficient}, where summation of vector indices is implied.

To get the gradient with respect the the signed coefficient values, the gradients of $f$ and $g$ are multiplied by the corresponding coefficient sign, i.e.
	\begin{align}
		\grad{\overline{c}}f & = \sign(\overline{c}) \times \grad{|\overline{c}|} f
\intertext{and}
		\grad{\overline{c}}g & = \sign(\overline{c}) \times \grad{|\overline{c}|} g
	\end{align}
where $\times$ indicates element-wise multiplication. The gradients with respect to the base, non-sorted set of wavelet coefficients $\{c\}$, $\grad{c}f$ and $\grad{c}g$ respectively, are found by performing the inverse sorting with respect to the absolute wavelet coefficient values. In this way $\grad{c}\mathcal{S}$ can be computed from $\grad{c} f$ and $\grad{c} g$ through Eq.~\eqref{eq:SparsityTermDerivative}. 

Having computed the gradient of the sparsity cost with respect to the output nodes (wavelet coefficients) we can now use standard back-propagation on the full network to compute the associated gradient on each entry in the low- and high-pass filter matrices. For a given, fixed filter length $N_{\mathrm{filt}}$, entries in the filter matrices which are identically zero are not modified by a gradient. Conversely, the gradient on every filter matrix entry to which a particular filter coefficient is contributing is added to the corresponding sparsity gradient in filter coefficient space, possibly with a sign change  in the case of high-pass filter matrices, cf.~Eq.~\eqref{eq:wavelet-coefficients-b}. In this way, the gradient on the wavelet coefficients is translated into a gradient in filter coefficient space, which we can then use in stochastic gradient descent, along with a similar regularisation gradient, to gradually improve our wavelet basis as parametrised by $\{a\}$.

\subsection{Regularisation term}

The regularisation terms are included to ensure that the optimal filter coefficient configuration does indeed correspond to an orthonormal wavelet basis as defined through conditions (C$1$--$5$). As noted in Section~\ref{subsec:theory-combining}, we choose to cast cast these conditions in the form of quadratic regularisation conditions on the filter coefficients $\{a\}$. Each of the conditions (C$1$--$5$) is of the form
	\begin{align}
		h_{k}(\{a\}) & = d_{k}
	\end{align}
which can be written as a quadratic regularisation term, i.e.
	\begin{align} \label{eq:regularisation-term-general}
		\mathcal{R}_{k}(\{a\}) & = \left( h_{k}(\{a\}) - d_{k} \right)^{2}
	\end{align}
and the combined regularisation term is then given by 
	\begin{align}
		\mathcal{R}(\{a\}) & = \sum_{k=1}^{5} \mathcal{R}_{k}(\{a\})
	\end{align}
This formulation allows for the search to proceed in the full $N_{\mathrm{filt}}$-dimensional search space, and the regularisation constant $\lambda$ regulates the degree  of precision to which the optimal filter coefficient configuration will fulfill conditions (C$1$--$5$).

In order to translate deviations from conditions (C$1$--$5$) into gradients in filter coefficient space, we take the derivative of each of the terms $\mathcal{R}_{k}$ with respect to the filter coefficients $a_{i}$. 
%
The gradients are found to be:
	\begin{align}
		\grad{a}\mathcal{R}_{1} & = \uvek{e}_{i} \, 2 \left(  \Big[ \sum_{k} a_{k} \Big] - \sqrt{2} \right) \tag{D$1$} \\		
		\grad{a}\mathcal{R}_{2} & = \uvek{e}_{i} \, \sum_{m} \, 2 \left( \sum_{k} \Big[ a_{k} a_{k+2m} \Big] - \delta_{m,0} \right) \nonumber \\
								 & \qquad\qquad \times \left(a_{i+2m} + a_{i-2m} \right) \tag{D$2$} \\
		\grad{a}\mathcal{R}_{3} & = \uvek{e}_{i} \, \sum_{m} \, 2 \left( \sum_{k} \Big[ b_{k} b_{k+2m} \Big] - \delta_{m,0} \right) \nonumber \\
								 & \qquad\qquad \times \left(a_{i+2m} + a_{i-2m} \right) \tag{D$3$} \\
		\grad{a}\mathcal{R}_{4} & = \uvek{e}_{i} \, 2 \left( \sum_{k} b_{k} \right) \times (-1)^{N - i - 1} \tag{D$4$} \\
		\grad{a}\mathcal{R}_{5} & = \vek{0} \tag{D$5$}
	\end{align}
Since condition (C$5$) is satisfied exactly by the definition in Eq.~\eqref{eq:wavelet-coefficients-b}, the corresponding gradient is identically equal to zero.

The combined gradient from the regularisation term is then the sum of the above five (four) contributions.

\section{Implementation} \label{sec:implementation}

The learning procedure based on the objective function and associated gradients presented in Section~\ref{sec:LearningProcedure} is implemented \cite{wavenet} as a publicly available \textsc{C++} \cite{Stoustrup1995} package. The matrix algebra operations are implemented using \textsc{armadillo} \cite{armadillo}, with optional interface to the high-energy physics \textsc{root} library \cite{root}.

This package allows for the processing of $1$- and $2$D dimensional training examples of arbitrary size, provides data generator for a few toy examples and reads CSV input  as well as high-energy physics collision events in the \textsc{HepMC} \cite{hepmc} format. The $2$D wavelet transform is perform by performing the $1$D transform on each row in the signal, concatenating the output rows, and then performing the $1$D transform on each of the resulting columns. Their matrix concatenation then corresponds to the $2$D set of wavelet coefficients.

In addition to standard (batch) gradient descent, the library allows for the use of gradient momentum and simulated annealing of the regularisation term in order to ensure faster and more robust convergence to the global minimum even in the presence of local minima and steep regularisation contours.

\section{Example: QCD $\mathbf{2 \to 2}$ processes in high-energy physics} \label{sec:example}

As an example of the procedure for learning optimal wavelet bases according to the metric presented in Section~\ref{sec:MeasuringSparsity}, using the implementation in Sections~\ref{sec:LearningProcedure} and \ref{sec:implementation}, we choose that of 
hadronic jets produced at proton colliders. In particular, the input to the training is taken to be simulated quantum chromodynamics (QCD) $2 \to 2$ processes, generated in \textsc{Pythia8} \cite{Sjostrand2006,Sjostrand2008}, segmented into a $2$D array of size $64 \times 64$ in the $\eta-\phi$ plane, roughly corresponding to the angular granularity of present-day general purpose particle detectors. The collision events are generated at a center of mass energy of $\sqrt{s} = 13~\mathrm{TeV}$ with a generator-level $p_{\perp}$ cut of $280~\mathrm{GeV}$ imposed on the leading parton.

QCD radiation patterns are governed by scale-independent splitting kernels \cite{Buckley2011}, which could make them suitable candidates for wavelet representation, since these naturally exhibit self-similar, scale-independent behaviour. In that case, the optimal (in the sense of Section~\ref{sec:MeasuringSparsity}) representation is one which efficiently encodes the localised angular structure of this type of process, and could be used to study, or even learn, such radiation patterns. In addition, differences in representation might help distinguish between such non-resonant, one-prong ``QCD jets'' and resonant, two-prong jets e.g.~from the hadronic decay of the $W$ and $Z$ eletroweak bosons.

We also note that, as alluded to in Section~\ref{subsec:theory-combining}, for signals of interest in collider physics, a standard neural network with ``wavenet'' architecture contains an enormous number of free parameters, e.g.~$N_{c}^{\mathrm{2D}} \approx 4.4\times 10^{7}$ for $N \times N = 64\times 64$ input, which is reduced to $N_{\mathrm{filt}}$, i.e.~as few as two, by the parametrisation in terms of the filter coefficients $\{a\}$.

\begin{figure}
	\centering%
	\includegraphics[width=0.45\textwidth]{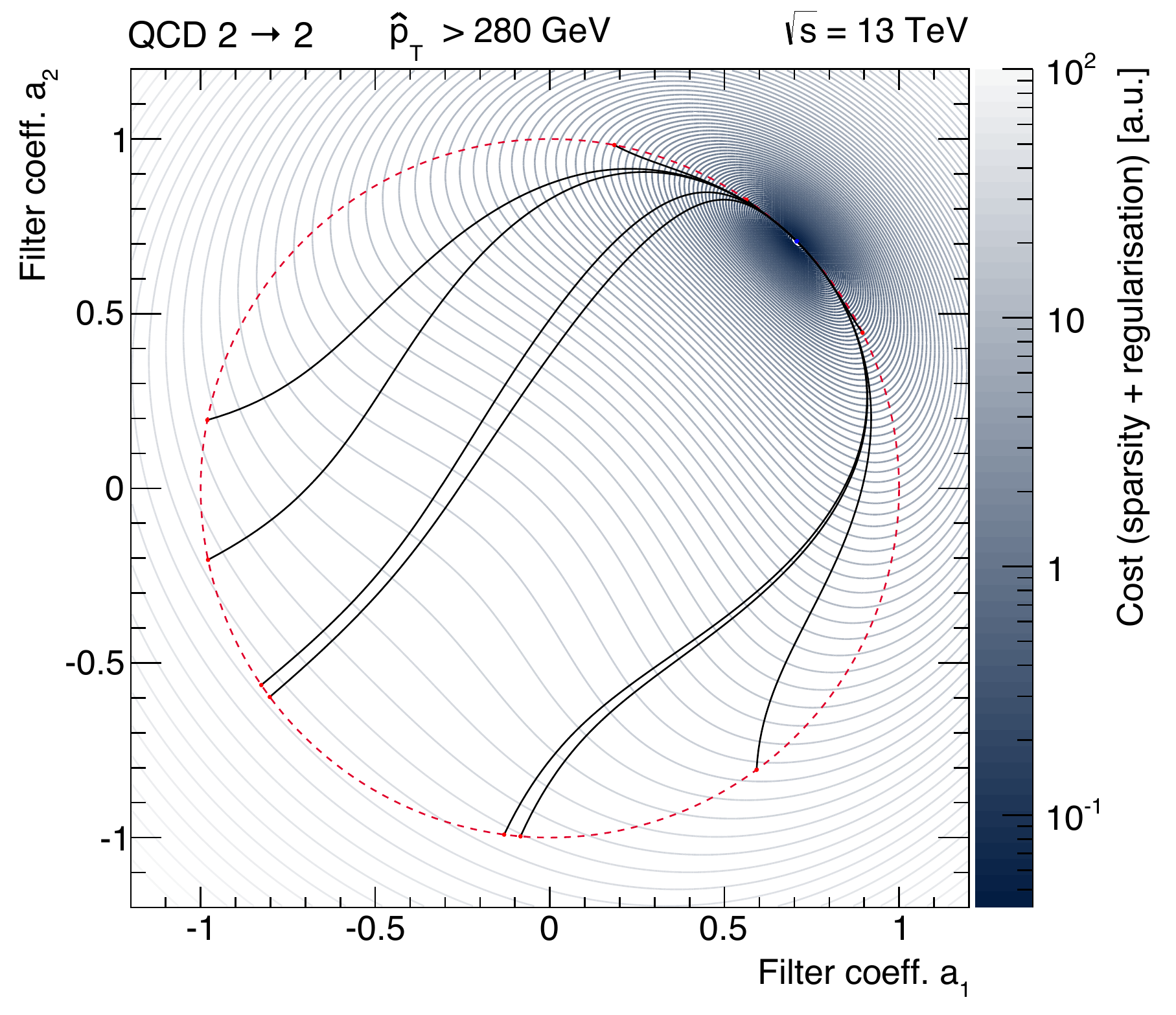}%
	\caption{Map of the average total cost (regularisation and sparsity) for QCD $2 \to 2$ events with $\hat{p}_{\perp} > 280 \unit{GeV}$, for only two filter coefficients $a_{1,2}$. Initial configurations are generated on the unit circle in the $a_{1}-a_{2}$ plane (red dots on dashed red line), to initially satisfy condition (C$2$), and better configurations are then learned iteratively (solid black lines) by using back-propagation with gradient descent, until a minimum (blue dot(s)) is found.}
	\label{fig:Wavelet-CostMap}
\end{figure}

We apply the learning procedure using Ref.~\cite{wavenet}, iterating over such ``dijet'' events pixelised in the $\eta-\phi$ plane, and use back-propagation with gradient descent to learn the configuration of $\{a\}$ which, for fixed $N_{\mathrm{filt}}$, minimises the combined sparsity and regularisation in Eq.~\eqref{eq:combined-cost-function}. This is shown in Fig.~\ref{fig:Wavelet-CostMap} for $N_{\mathrm{filt}} = 2$. 

It is seen that, for $N_{\mathrm{filt}} = 2$, only one minimum exists, due to only one point in $a_{1}-a_{2}$ space fulfilling all five conditions (C$1$--$5$). This configuration has $a_{1} = a_{2} = 1/\sqrt{2}$ and is exactly the Haar wavelet \cite{Haar1919}. Although this is an instructive example allowing for clean visualisation, showing the clear effect of the gradient descent algorithm and the efficacy of the interpretation of conditions (C$1$--$5$) as quadratic regularisation terms, it also doesn't tell us much since the global minimum will be the same for all classes of inputs. For $N_{\mathrm{filt}} > 2$ the regularisation allows for minima in an effective hyperspace with dimension $D > 0$.

\begin{figure*}
	\centering%
	\begin{subfigure}[b]{0.33\textwidth}
		\centering%
		\captionsetup{justification=centering}%
		\includegraphics[width=0.9\textwidth]{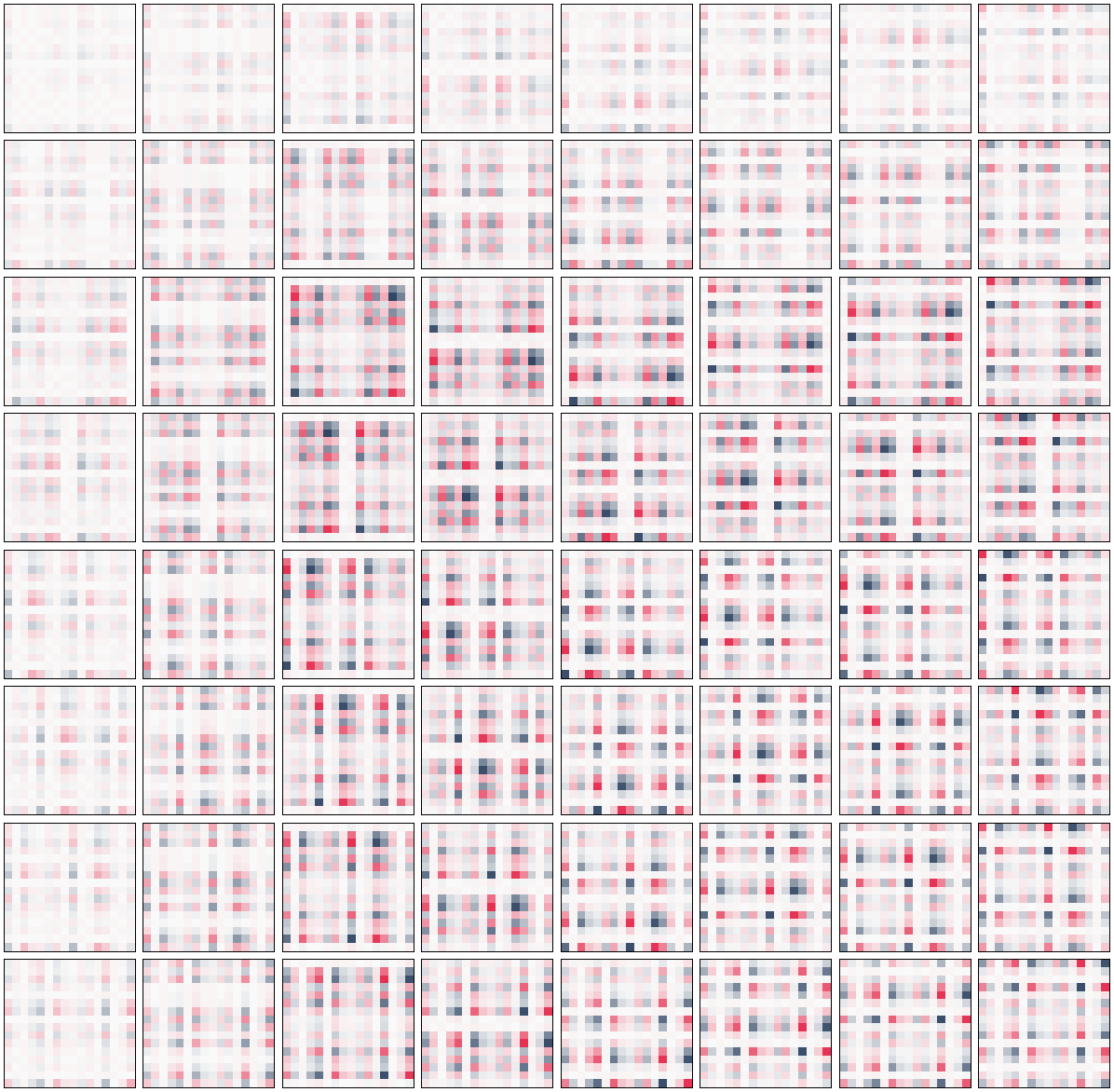}
		\caption{Initial configuration}
		\label{subfig:exampleBases1}
	\end{subfigure}%
	\begin{subfigure}[b]{0.33\textwidth}
		\centering%
		\captionsetup{justification=centering}%
		\includegraphics[width=0.9\textwidth]{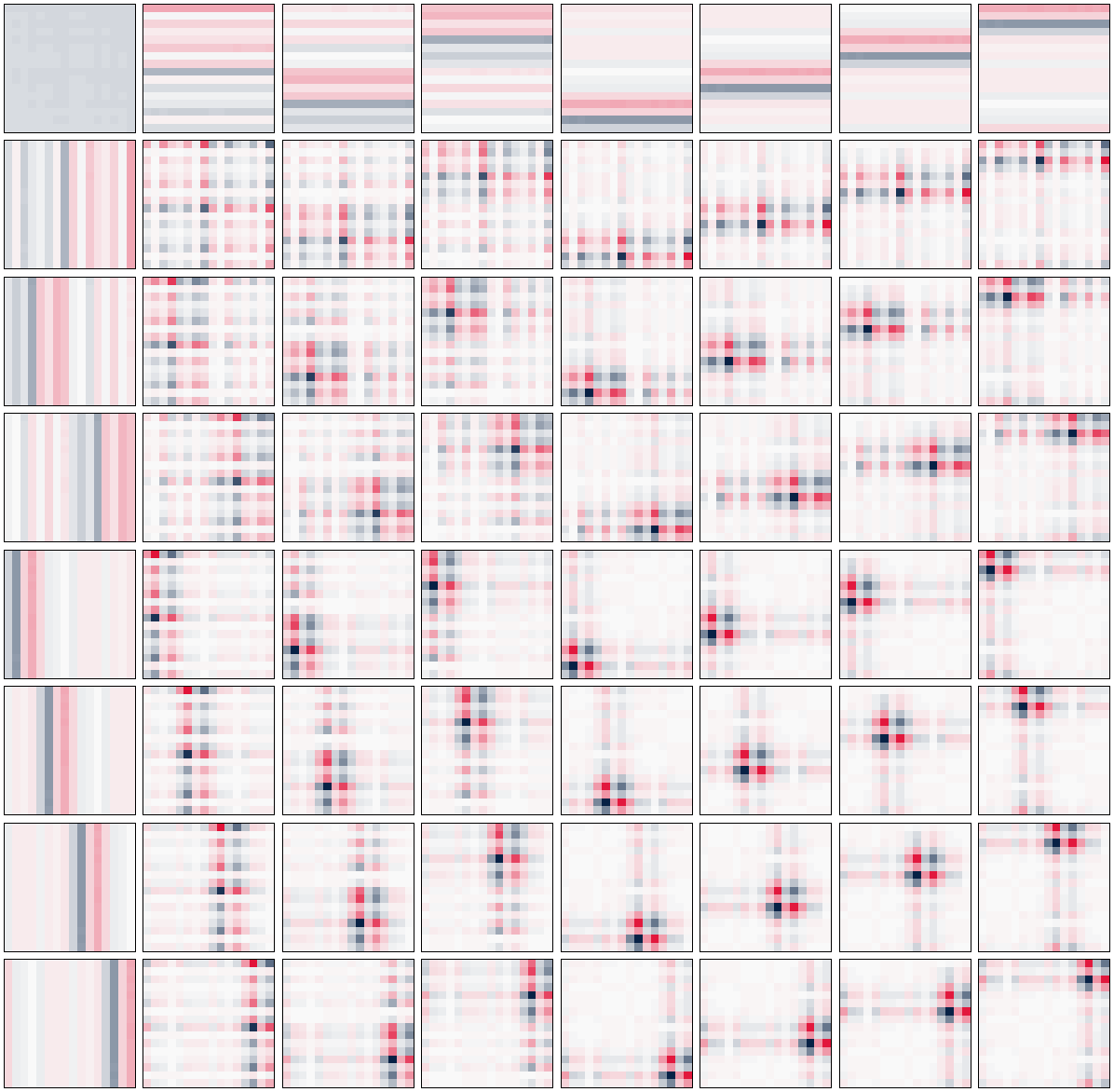}
		\caption{Intermediate configuration}
		\label{subfig:exampleBases2}
	\end{subfigure}%
	\begin{subfigure}[b]{0.33\textwidth}
		\centering%
		\captionsetup{justification=centering}%
		\includegraphics[width=0.9\textwidth]{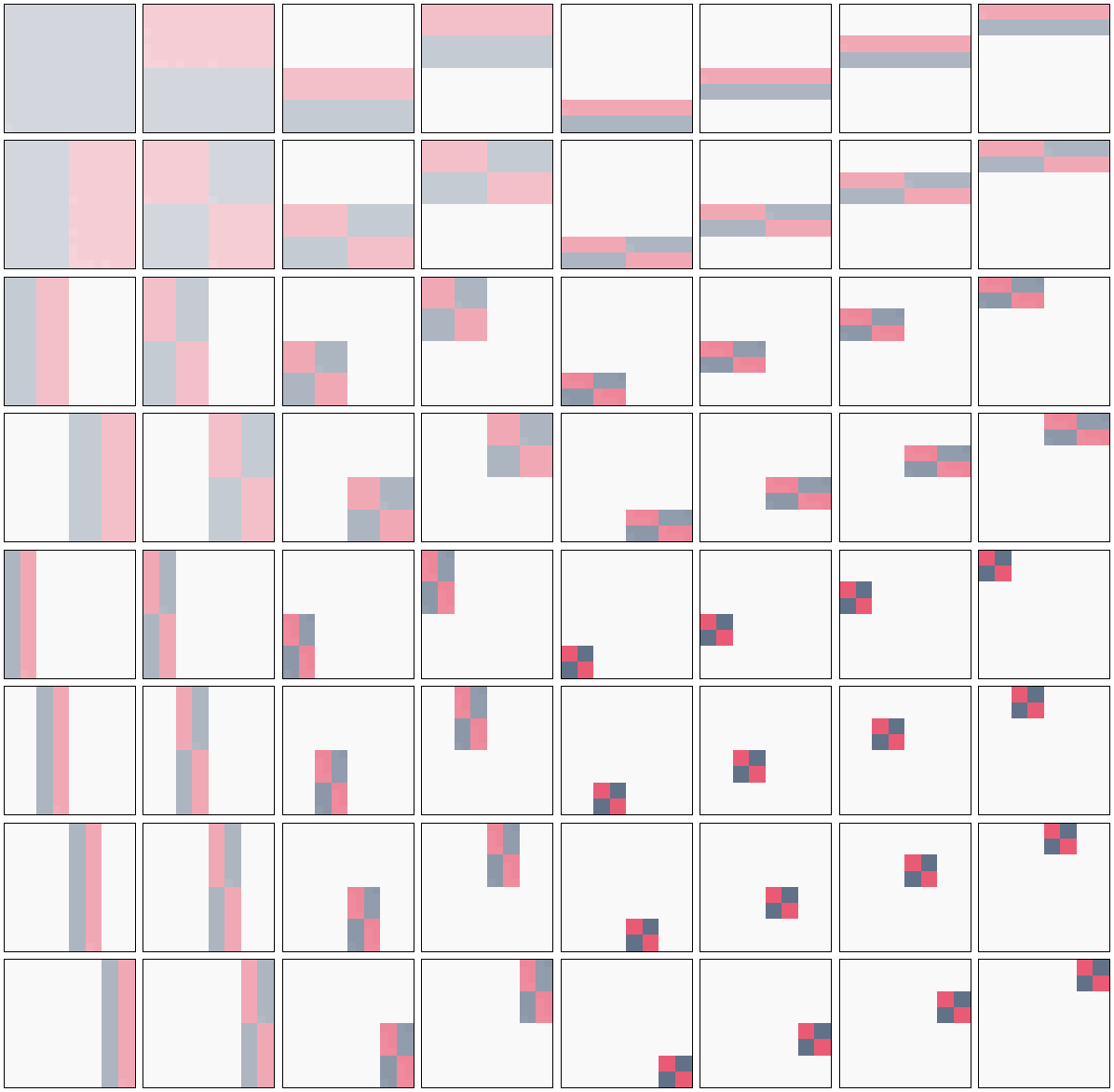}
		\caption{Final configuration}
		\label{subfig:exampleBases3}
	\end{subfigure}%
	\caption{Examples of the $64$ lowest-scale $2$D wavelet basis functions, found by optimisation on rasterised QCD $2 \to 2$ events with $\hat{p}_{\perp} > 280 \unit{GeV}$ in an $N_{\mathrm{filt}} = 16$-dimensional filter coefficient space, (a) at initialisation, (b) at an intermediate point during training and (c) at termination of the learning procedure upon convergence.}
	\label{fig:exampleBases}
\end{figure*}

Instead choosing $N_{\mathrm{filt}} = 16$ we can perform the same optimisation, but now with sufficient capacity of the wavelet basis to encode the defining features of this class of signals. The effect of the learning procedure is presented in Figure~\ref{fig:exampleBases}, showing a selection of the lowest-scale wavelet basis functions corresponding to particular filter coefficient configurations at the beginning of, during, and at convergence of the learning procedure in this higher-dimensional search space.

The random initialisation on the unit hyper-sphere is shown to produce random noise (Figure~\ref{subfig:exampleBases1}), which does not correspond to a wavelet basis, since the algorithm has not yet been afforded time to update the filter coefficients to conform with the regularisation requirements. At some point roughly half way through the training, the filter coefficient configuration does indeed yield an orthonormal wavelet basis (Figure~\ref{subfig:exampleBases2}), and the learning procedure now follows the gradients towards greater sparsity along a high-dimensional, quadratic regularisation ``valley''. Finally, at convergence, the optimal wavelet found is again seen to be exactly the Haar wavelet (Figure~\ref{subfig:exampleBases3}), despite the vast amount of freedom provided the algorithm by virtue of $16$ filter coefficients. That is, the learning procedure arrives at the optimal configuration by setting $14$ filter coefficients to exactly zero without any manual tuning. 

This result shows that limiting the support of the basis functions provides for more efficient representation than any deviations due to radiation patterns could compensate for. Indeed, it can be show that removing some of the conditions (C$1$--$5$) so as to ensure that $\{a\}$ simply corresponds to an orthonormal basis (i.e.~not necessarily an orthonormal \emph{wavelet} basis) the learning procedure results in \emph{the pixel basis}, i.e.~the one in which each basis function corresponds to a single entry in the input array. This shows that, due to the fact that QCD showers are fundamentally point-like (due to the constituent particles) and since they, to leading order, are dominated by a few particles carrying the majority of the energy in the jet, the representation which best allows for representation of single particles will prove optimal according to our chosen measure Eq.~\eqref{eq:GiniCoefficient}. However, since this example studies the optimal representation of entire event, its conclusions may change for inputs restricted to a certain region in $\eta-\phi$ space around a particular jet, i.e.~for the study of optimal representation of jets themselves.


% ...

\section*{Acknowledgments}

The author is supported by the Scottish Universities Physics Alliance (SUPA) Prize Studentship. The author would like to thank Troels C.~Petersen for insightful discussions on the subject matter, and James W.~Monk for providing Monte Carlo samples.


\bibliographystyle{elsarticle-num}
\bibliography{./Bibliography}

\end{document}